\documentclass[10pt,twocolumn,letterpaper]{article}

\PassOptionsToPackage{table}{xcolor}
\usepackage[pagenumbers]{cvpr}      %

\usepackage{multirow}
\usepackage{microtype}
\usepackage{amsthm}

\AtEndPreamble{%
  \crefname{proposition}{Prop.}{Props.}\Crefname{proposition}{Proposition}{Propositions}%
  \crefname{equation}{Eq.}{Eqs.}%
  \crefname{figure}{Fig.}{Figs.}\Crefname{figure}{Figure}{Figures}}
\usepackage{amsmath,amssymb}
\newcommand{\ALC}{\textsc{ALC}}
\newcommand{\ClockU}{\textsc{Uniform-FT}}
\newcommand{\Uni}{\textsc{Uniform}}

\definecolor{revpurple}{RGB}{128,0,170}
\newcommand{\rev}[1]{#1}

\definecolor{revpink}{RGB}{230,0,126}
\newcommand{\revp}[1]{#1}

\definecolor{revbrown}{RGB}{150,80,10}

\definecolor{revteal}{RGB}{0,128,128}
\newcommand{\revt}[1]{#1}

\definecolor{revblue}{RGB}{0,70,230}

\definecolor{revorange}{RGB}{225,105,0}

\newcommand{\revm}[1]{#1}
\newcommand{\revmon}{}
\definecolor{revgreen}{RGB}{0,120,60}
\newcommand{\revk}[1]{#1}

\definecolor{revred}{RGB}{200,30,30}
\newcommand{\revr}[1]{#1}

\definecolor{revfpurple}{RGB}{128,0,170}
\newcommand{\revf}[1]{#1}

\newcommand{\revw}[1]{#1}

\newcommand{\revi}[1]{#1}

\newcommand{\revj}[1]{#1}
\newcommand{\revl}[1]{#1}
\newcommand{\revs}[1]{#1}
\newcommand{\revc}[1]{#1}

\newcommand{\colfit}[1]{\vspace{#1}}
\definecolor{cvprblue}{rgb}{0.21,0.49,0.74}
\definecolor{lcBest}{HTML}{C7DCF1}
\definecolor{lcSecond}{HTML}{E8F1FA}

\usepackage{tikz}
\usetikzlibrary{arrows.meta}
\definecolor{lcGroup}{HTML}{ECEEF1}%
\definecolor{lcOursRow}{HTML}{E6F0FA}%
\definecolor{lcMain}{HTML}{F6F0DF}%
\definecolor{lcGain}{HTML}{1E7B34}%
\definecolor{lcIcon}{HTML}{2563AE}%
\definecolor{lcIconB}{HTML}{DC5A2E}%
\definecolor{lcIconG}{HTML}{8D939B}%
\newcommand{\sd}[1]{{\fontsize{5.6}{6}\selectfont\,$\pm$#1}}%
\newcommand{\bst}[1]{\textbf{#1}}%
\newcommand{\snd}[1]{\underline{#1}}%
\newcommand{\gain}[1]{\textcolor{lcGain}{\textbf{#1}}}
\newcommand{\up}{\,$\uparrow$}
\newcommand{\dn}{\,$\downarrow$}
\newcommand{\toprulex}{\specialrule{0.9pt}{0pt}{2pt}}
\newcommand{\bottomrulex}{\specialrule{0.9pt}{1pt}{0pt}}
\newcommand{\tgroup}[2]{\rowcolor{lcGroup}\multicolumn{#1}{@{}l}{\textbf{\textit{#2}}}\\}
\newcommand{\venue}[1]{\,{\fontsize{5.6}{6}\selectfont\textcolor{black!55}{#1}}}

\usepackage{trimclip}
\newsavebox{\lcfitbox}
\newif\iflctight\lctightfalse
\newdimen\lcminsep\lcminsep=2pt
\newdimen\lcnewsep
\newcommand{\lcfit}[3]{\sbox{\lcfitbox}{#3}%
  \wlog{LCFIT natural \the\wd\lcfitbox\space target \the\dimexpr#1\relax}%
  \ifdim\wd\lcfitbox>#1\relax
    \iflctight
      \lcnewsep=\dimexpr\tabcolsep-(\wd\lcfitbox-#1)/(2*#2-2)\relax
      \ifdim\lcnewsep<\lcminsep \ifdim\tabcolsep>\lcminsep\lcnewsep=\lcminsep\else\lcnewsep=\tabcolsep\fi\fi
      {\setlength{\tabcolsep}{\lcnewsep}\sbox{\lcfitbox}{#3}%
       \wlog{LCFIT tight \the\wd\lcfitbox\space target \the\dimexpr#1\relax}%
       \ifdim\wd\lcfitbox>\dimexpr#1+0.5pt\relax\resizebox{#1}{!}{\clipbox{0pt 0pt 0pt 0pt}{\usebox{\lcfitbox}}}%
       \else\makebox[#1][c]{\clipbox{0pt 0pt 0pt 0pt}{\usebox{\lcfitbox}}}\fi}%
    \else\resizebox{#1}{!}{\clipbox{0pt 0pt 0pt 0pt}{\usebox{\lcfitbox}}}\fi
  \else{\setlength{\tabcolsep}{\dimexpr\tabcolsep+(#1-\wd\lcfitbox)/(2*#2-2)\relax}\sbox{\lcfitbox}{#3}\clipbox{0pt 0pt 0pt 0pt}{\usebox{\lcfitbox}}}\fi}
\renewcommand{\arraystretch}{1.18}
\usepackage[pagebackref,breaklinks,colorlinks,allcolors=cvprblue]{hyperref}
\hypersetup{pdftitle={The Latent Clock: Predicting and Repairing Collision Failure in Video Generation}}%

\def\confName{CVPR}
\def\confYear{2027}

\title{The Latent Clock: Predicting and Repairing Collision Failure in Video Generation}
\author{%
\begin{minipage}{\dimexpr\textwidth-2\tabcolsep\relax}
\centering
\normalsize
Taoran~Fu\textsuperscript{1,*,\dag}\hspace{0.55em}\ignorespaces Ziqian~Wang\textsuperscript{2,*}\hspace{0.55em}\ignorespaces Haoyang~Wu\textsuperscript{3,*}\hspace{0.55em}\ignorespaces
Huanyao~Zhang\textsuperscript{4}\hspace{0.55em}\ignorespaces
Ji~Jiang\textsuperscript{1}\hspace{0.55em}\ignorespaces Jiepeng~Zhou\textsuperscript{5}\hspace{0.55em}\ignorespaces Long~Bai\textsuperscript{1}\hspace{0.55em}\ignorespaces
Qinghua~Zhao\textsuperscript{6,\ddag}\\[4pt]
\small
\textsuperscript{1}Xi'an Jiaotong-Liverpool University\quad\textsuperscript{2}Beijing Institute of Technology\quad\textsuperscript{3}Huazhong University of Science and Technology\\\textsuperscript{4}Peking University\quad\textsuperscript{5}The Hong Kong University of Science and Technology (Guangzhou)\quad\textsuperscript{6}Hefei University\\\textsuperscript{*}Equal contribution\quad\textsuperscript{\dag}Project lead: taoran.fu26@student.xjtlu.edu.cn\quad\textsuperscript{\ddag}Corresponding author: zhaoqh@buaa.edu.cn\vspace{-13pt}
\end{minipage}%
}

\usepackage{etoolbox}\makeatletter\patchcmd{\@maketitle}{\vskip .375in}{\vskip 0in}{}{\PackageError{arxiv}{title patch failed}{}}\makeatother
\begin{document}

\maketitle

\begin{abstract}
\revr{Latent video generation models apply temporal compression, which encodes each group of consecutive frames into a single latent representation. Wan2.2-TI2V-5B and LTX-Video, two such models, fail to generate an accurate collision in all 224 test cases of a falling ball. To study collision failure, we build LatentClock-Bench, about 120{,}000 simulated videos of a ball hitting the ground, a wall, or another ball, each with a known collision time, ball speed, and ball size. For our models trained on LatentClock-Bench, collision failure rises with the distance that the ball moves within one group relative to the ball size, a distance that grows with the ball speed and the number of frames per group. Collision failure also varies with where the collision falls within a group and repeats one group later. When existing models encode and decode these videos, up to 58.3\% of collisions fail, and 64\% of all failed collisions have a wrong speed after the collision. Based on these findings, we propose \emph{Adaptive Latent Clock (\ALC)}, which keeps the number of frames per group fixed, concentrates frames around each collision time to shorten this distance, and supervises the speed after the collision in encoder and decoder training. Experiments show that collision failure falls from \revc{28.1\%} to \revc{4.8\%} on \mbox{LatentClock-Bench} and from 121 to 30 of 152 videos regenerated by LTX-Video.}
\end{abstract}

\suppressfloats[t]
\begin{figure}[t]
\centering
\includegraphics[width=1.0\columnwidth]{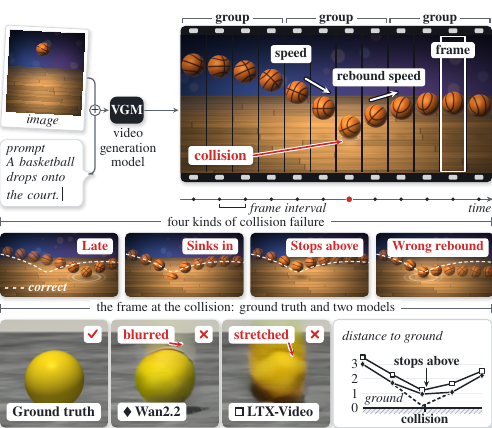}
\caption{\revl{\textbf{Collision failure in generated videos.} Frames form groups, and a generated collision can fail in four ways.}}
\label{fig:teaser}
\end{figure}

\section{Introduction}\label{sec:intro}

\revj{A world model has to reproduce collisions, among the most common interactions between objects, at the right time and with the right rebound speed. Video generation models are increasingly studied as such world models~\cite{brooks2024sora,nvidia2025cosmos}. Most current video generation models~\cite{brooks2024sora,yang2025cogvideox,nvidia2025cosmos} use a diffusion transformer as the generator to produce latent representations (\emph{latents}) and a video tokenizer, which converts between frames and latents, to decode them into frames~\cite{rombach2022ldm,peebles2023dit,yu2024magvitv2,zhao2024cvvae}. Many video tokenizers, including those of Wan, HunyuanVideo and LTX-Video, apply \emph{temporal compression}, which represents each group of four to eight consecutive frames with a single latent~\cite{wan2025wan,kong2024hunyuanvideo,hacohen2024ltx}. A collision lasts a few milliseconds, so the speed of an object changes between two consecutive frames, whereas one latent spans 167 to 500\,ms.}\looseness=-1

\revj{Existing work on temporal compression develops video tokenizers in two directions. Fixed-rate tokenizers represent every group of four to eight consecutive frames with one latent and aim to reduce computation, to decode fast motion more accurately, or to make latents easier to generate~\cite{li2025wfvae,cheng2025leanvae,xing2025videovaeplus,argaw2025causalvae,skorokhodov2025diffusability,liu2025ssvae}. Adaptive tokenizers vary the temporal compression or the number of tokens across a video according to motion, duration or information content, mainly to reduce computation further~\cite{yuan2025dlfrvae,zhong2025vfrtok,li2026adaptok,xiong2026evatok,yan2025elastictok}. Both directions evaluate tokenizers by computation and by reconstruction or generation quality averaged over whole videos, and neither evaluates collisions specifically. However, current models with temporal compression fail at collisions. \cref{fig:teaser} illustrates such a failure for a falling ball and compares the videos regenerated by Wan2.2-TI2V-5B~\cite{wan2025wan} and LTX-Video~\cite{hacohen2024ltx} with the simulated ground truth frame by frame and through the distance between the ball and the ground. During the fall, both generated balls stay close to the ground truth. At the collision, however, the ball generated by Wan2.2-TI2V-5B is blurred, the ball generated by LTX-Video is stretched, and both stop about one ball radius above the ground. The failure is not limited to this case: given 224 first frames of a ball released above the ground, neither model generates a single collision at the expected time and with the expected rebound speed.}\looseness=-1

\revj{To study how temporal compression and the properties of a collision affect these failures, we build LatentClock-Bench, about 120,000 videos simulated with MuJoCo~\cite{todorov2012mujoco}, in which a ball is dropped onto the ground, thrown against a wall or sent into a second ball under varied gravity, elasticity, ball size and initial state. Since the videos are simulated, the quantities that define each collision are known, including the position and diameter of the ball in each frame, the frame rate, the collision time and the speeds before and after the collision. These known quantities allow each collision to be verified: we track the ball in a video and compare its collision time and speeds with the simulated values. In a latent model, a collision can fail in the tokenizer, which converts between frames and latents, or in the generator, which produces the latents. We examine both. We encode and decode the benchmark videos with the tokenizers of eight current models, and train generators on the benchmark with 1, 2, 4 and 8 frames per group while varying the speed of the ball and the collision time. In the generators, collision failure rises with the number of frames per group and with the distance that the ball travels across one group of frames relative to its size, and it repeats with a period of one group as the collision time shifts. In the tokenizers, encoding and decoding break many collisions, mostly through a wrong rebound speed. These results point to two targets for repair: the grouping of frames around the collision time and the rebound speed in the tokenizer.}\looseness=-1

\revj{Based on these findings, we propose \emph{Adaptive Latent Clock} (\ALC). We call the times at which the frames sample the scene the \emph{latent clock}, since they set the time that each latent covers. To shorten the distance that the ball travels within the group containing a collision, \ALC{} estimates each collision time, concentrates frames around it at fixed temporal compression and passes the resulting frame times to the generator. To correct the rebound speed that the tokenizers break, it further supervises the speed of the ball around the collision in tokenizer training.}\looseness=-1

\revj{Experiments on LatentClock-Bench demonstrate that \ALC{} lowers collision failure from \revc{28.1\%} to \revc{4.8\%} at the same temporal compression. Removing Frame Timing or Rebound Supervision raises it to \revc{7.0\%} or \revc{10.7\%}, respectively. On LTX-Video, \ALC{} also lowers the collision failures in regenerated videos from 121 to 30 of 152.}\looseness=-1

\revj{In summary, we introduce LatentClock-Bench, a simulated benchmark with known collision quantities, and show that collision failure grows with the number of frames per group and the ball's travel within a group and originates in part from the tokenizer. We further propose \ALC{}, which lowers this failure at fixed temporal compression by concentrating frames around collisions and constraining the rebound speed in the tokenizer. We will release the benchmark, code and models.}\looseness=-1

\section{Related Work}\label{sec:related}

\begin{figure*}[t]
  \centering
  \includegraphics[width=\textwidth]{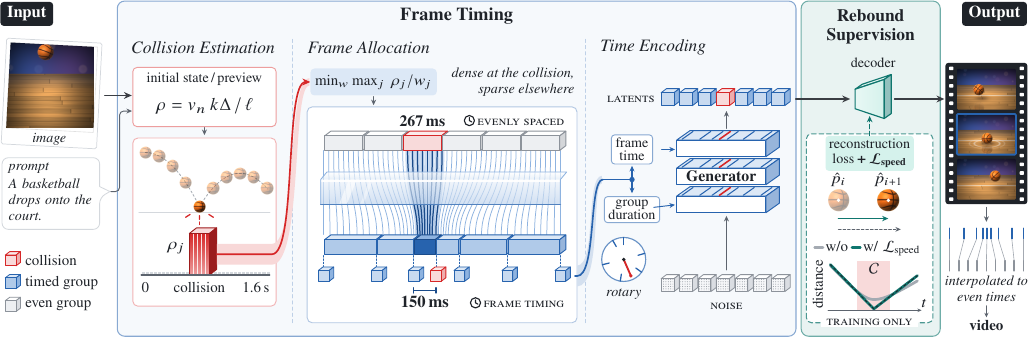}
  \caption{\textbf{Overview of \ALC.} Frame Timing shortens the groups of frames around each collision, and the generator reads the resulting frame times. Rebound Supervision supervises the speed of the ball around the collision when the tokenizer is trained.}
\label{fig:alc_framework}
\end{figure*}

\looseness=-1 Recent work studies video physics through benchmarks, \rev{analyses} and interventions. Benchmarks compare continuations with experiments~\cite{motamed2025physicsiq,radsch2026verified}, score physical commonsense~\cite{bansal2025videophy,bansal2025videophy2,meng2024phygenbench} and laws~\cite{kang2025howfar}, and test object permanence~\cite{zhang2026wrop}. Analyses report unstable frame rates~\cite{gao2026pulse} and motion slower than Earth gravity~\cite{thozhiyoor2025slower}, or trace violations to attention~\cite{li2026violate} and decode physics from denoiser states~\cite{esmati2026invisible}. Interventions steer sampling via time control~\cite{wu2026seeing,wang2026bullettime,huang2026spacetimepilot} or prompts~\cite{xue2025phyt2v,wang2026chaincot}, inject simulation~\cite{foo2026psivg,liu2024physgen} or physics guidance~\cite{zhang2025videorepa,wang2025wisa,pathak2026physvid,shen2026phantom,yuan2026itpa}, or optimize force goals~\cite{gillman2026goalforce} and rewards~\cite{zhang2026physrvg}, or post-train on falling objects~\cite{li2025pisa}. The Seriality Gap~\cite{diazchao2026seriality} traces failures on long chains of hard-sphere events to too little serial computation, whereas we trace failures of single collisions to temporal compression.

\revl{Most video generation models operate in the latent space of a video tokenizer~\cite{rombach2022ldm,yu2024magvitv2,zhao2024cvvae}. Fixed-rate tokenizers represent each group of four to eight consecutive frames with one latent~\cite{yang2025cogvideox,kong2024hunyuanvideo,wan2025wan,hacohen2024ltx,nvidia2025cosmos,genmo2024mochi,hacohen2026ltx2,hunyuan2025video15}. Recent ones reduce computation~\cite{li2025wfvae,cheng2025leanvae}, decode fast motion more accurately~\cite{xing2025videovaeplus,argaw2025causalvae}, or make latents easier to generate~\cite{skorokhodov2025diffusability,liu2025ssvae}. IV-VAE~\cite{wu2025ivvae} reports uneven reconstruction quality within a group.}\looseness=-1

\revl{Adaptive methods vary the number of frames or tokens that represent each part of a video. They follow motion~\cite{yuan2025dlfrvae,yuan2025dlfrgen,zheng2026dvg}, duration~\cite{zhong2025vfrtok}, compression~\cite{dong2026mtcvae}, quality or information content~\cite{li2026adaptok,xiong2026evatok,yan2025elastictok,ye2026infotok}, uniqueness~\cite{yuan2026unicomp}, and surprise~\cite{tang2026surge}. Other designs order tokens from coarse to fine~\cite{atanov2026videoflextok}. These methods follow content signals mainly to reduce computation, so a brief collision receives no more frames than the motion around it. EVD~\cite{maduabuchi2026evd} gates denoising to repair interaction failures. DVG~\cite{zheng2026dvg} also previews cheap denoising steps, but changes the token count to save computation, whereas \ALC{} fixes the latent count and places latents at collisions. Our experiments place four of these signals in our frame allocation. Continuous time encodings specify when frames occur~\cite{wang2026bullettime,zhong2025vfrtok}, and ArbInterp~\cite{zhang2026arbinterp} conditions interpolation on frame times without temporal compression. \ALC{} keeps temporal compression fixed, concentrates frames around collisions, and passes their frame times to the generator with the time encoding of BulletTime~\cite{wang2026bullettime}.}

\section{LatentClock-Bench}\label{sec:benchmark}\label{sec:construction}

\revl{LatentClock-Bench is designed to examine how temporal compression and the properties of a collision affect collision failure. For this purpose, each video should contain one collision whose time and speeds are known, the collision should be movable to any frame within a group of frames, and its outcome should be measurable from the frames. Existing benchmarks do not allow a collision to be moved within a group of frames~\cite{bansal2025videophy2,kang2025howfar,motamed2025physicsiq,tragoudaras2026morpheus}, as shown in \cref{tab:benchcmp}. LatentClock-Bench simulates each collision and shifts it frame by frame within a group while the scene stays fixed.}

\begin{table}[t]
\centering\small
\setlength{\tabcolsep}{4pt}\renewcommand{\arraystretch}{0.9}
\lcfit{\columnwidth}{5}{%
\begin{tabular}{@{}lcccc@{}}
\toprulex
\textbf{Benchmark} & Known & Movable & Measurable & Multiple fps \\
\midrule
VideoPhy-2~\cite{bansal2025videophy2} & $\times$ & $\times$ & $\times$ & $\times$ \\
Physics-IQ~\cite{motamed2025physicsiq} & $\times$ & $\times$ & $\times$ & $\times$ \\
Kang et al.~\cite{kang2025howfar} & $\checkmark$ & $\times$ & $\times$ & $\times$ \\
Morpheus~\cite{tragoudaras2026morpheus} & $\times$ & $\times$ & $\times$ & $\times$ \\
Seriality Gap~\cite{diazchao2026seriality} & $\checkmark$ & $\times$ & $\times$ & $\times$ \\
\midrule
LatentClock-Bench & $\checkmark$ & $\checkmark$ & $\checkmark$ & $\checkmark$ \\
\bottomrulex
\end{tabular}}
\caption{\textbf{Benchmarks for physics in video generation.} Existing benchmarks do not move a collision within a group of frames.}
\label{tab:benchcmp}\colfit{-7pt}%
\end{table}

\revs{We generate the videos of LatentClock-Bench by simulation with MuJoCo~\cite{todorov2012mujoco} under a fixed random seed. Each scene is of one of three types. In the first type, a ball released above the ground falls, hits the ground and bounces back up, as in \cref{fig:teaser}. In the second type, a ball thrown toward a vertical wall hits the wall and rebounds from it. In the third type, a moving ball hits a second ball, and the two balls move apart after the collision. For each scene, we sample the gravity, the elasticity of the collision, which is the ratio of the speed of the ball after the collision to that before it and lies between 0.55 and 0.85, the size of the ball, with a radius of 5, 8 or 12 pixels, and its initial state, which is the position and velocity of the ball at the start of the scene. We simulate each scene for 1.6\,s with one collision, whose contact lasts about 4\,ms, and record the position of the ball throughout the scene, the collision time and the speeds before and after the collision, which we call the \emph{collision speed} and the \emph{rebound speed}.}\looseness=-1

\revs{We then render each simulated scene at 120 frames per second (fps) with frames of $64\times64$ pixels and keep every sixth, fourth or second rendered frame, which gives videos at 20, 30 or 60\,fps, with 49 frames at 30\,fps. For part of the scenes, we further make copies in which the collision occurs one frame later at a time while the motion of the ball is otherwise unchanged. In total, LatentClock-Bench contains 122{,}500 videos, each provided with its frame rate, the position of the ball in each frame, the two consecutive frames between which the collision occurs, the collision time, the speeds of the ball before and after the collision, and the sampled gravity, elasticity and ball size.}\looseness=-1
\section{\revs{Failure Analysis on LatentClock-Bench}}\label{sec:analysis}

\revs{The videos of LatentClock-Bench are simulated, so their collisions are correct and their records serve as the reference. We give each model a simulated video or its first frame, compare the collision in the output video with the records, and determine under which conditions and in which part of the model collisions fail.}\looseness=-1

\begin{figure*}[t]
  \centering
  \includegraphics[width=\textwidth]{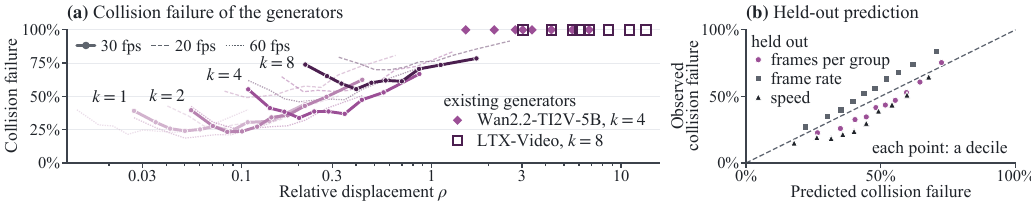}
  \caption{\textbf{Collision failure against the relative displacement.} (a) Collision failure of our generators and of two existing generators. (b) Observed collision failure against that predicted from the relative displacement when one factor is held out.}
\label{fig:law}\colfit{-2.3pt}
\end{figure*}

\subsection{\revs{Models and Measurement}}\label{sec:models}\label{sec:metrics}

\revs{We examine collisions at two stages. At the tokenizer stage, the tokenizer encodes and decodes a complete simulated video without the generator. At the generation stage, the tokenizer and the generator together produce a video from its first frame.}\looseness=-1

\revs{Among existing models, we examine eight tokenizers, those of Wan2.1 and Wan2.2~\cite{wan2025wan}, HunyuanVideo~\cite{kong2024hunyuanvideo}, CogVideoX-2b~\cite{yang2025cogvideox}, Cosmos CV4 and CV8~\cite{nvidia2025cosmos}, Mochi~\cite{genmo2024mochi} and LTX-Video~\cite{hacohen2024ltx}, on videos rendered at the frame rate and resolution that each tokenizer expects. We also examine the generators of Wan2.2-TI2V-5B and LTX-Video on the first frames of 224 videos in which a ball falls onto the ground from rest under Earth gravity. We restrict this test to falling balls because a single first frame shows no initial velocity, and only a ball starting from rest has a motion that its first frame determines. The existing models differ in architecture, training data and number of frames per group at the same time. To change the number of frames per group alone, we also train our own tokenizer and generator, a diffusion transformer with 33M parameters, on the training videos of the benchmark with 1, 2, 4 and 8 frames per group.}\looseness=-1

\begin{table}[t]
\centering\small
\setlength{\tabcolsep}{3.2pt}\renewcommand{\arraystretch}{0.95}
\lcfit{\columnwidth}{6}{%
\begin{tabular}{@{}lccccc@{}}
\toprulex
 & & \multicolumn{2}{c}{\textbf{PSNR} (dB)\,\up} & \multicolumn{2}{c}{\textbf{Failure} (\%)\,\dn} \\
\cmidrule(lr){3-4}\cmidrule(l){5-6}
\textbf{Tokenizer} & $k$ & UCF-101 & Ours & collision & geometric \\
\midrule
Wan2.1 & 4 & 35.3 & 49.7 & 13.0 & 2.6 \\
Wan2.2 & 4 & 35.7 & 53.6 & 2.7 & 2.2 \\
HunyuanVideo & 4 & 36.7 & 51.2 & 4.7 & 4.2 \\
CogVideoX-2b & 4 & 35.5 & 48.6 & 6.9 & 1.9 \\
Cosmos CV4 & 4 & 33.6 & 40.5 & 30.2 & 7.9 \\
Mochi & 6 & 31.6 & 39.8 & 32.4 & 9.4 \\
Cosmos CV8 & 8 & 31.9 & 43.3 & 58.3 & 15.2 \\
LTX-Video & 8 & 32.1 & 43.7 & 16.5 & 1.3 \\
\bottomrulex
\end{tabular}}
\caption{\textbf{Eight existing tokenizers.} PSNR and failure of the videos that each tokenizer encodes and decodes.}
\label{tab:clockcard}\colfit{4pt}%
\end{table}

\revs{We track the ball in each frame of an output video by fitting its profile with the known radius, locate the collision from its distance to the surface that it hits, and estimate its speeds before and after the collision. A collision is accurate when it meets three conditions: it is at most 0.1\,s early or late, the ball neither sinks into the surface by more than 10\% of its radius nor stops more than 15\% above it, and the ratio of rebound to collision speed stays within a speed-dependent tolerance of the simulated value. \emph{Collision failure} is the percentage of collisions that are not accurate, including those that cannot be measured. \emph{Geometric failure} is the percentage of collisions that violate the first two conditions, and \emph{rebound failure} is the percentage of the collisions meeting the first two conditions that violate the third. To check the measurement itself, we apply it to the simulated test videos, whose collisions are accurate. It accepts 5{,}365 of their 5{,}382 collisions, rejects none and leaves 17 unmeasured, and it detects 98.3\% to 100\% of the errors that we inject into these videos.}\looseness=-1

\subsection{\revs{Sources of Collision Failure}}\label{sec:sources}\label{sec:phenomenon}\label{sec:predictive}\label{sec:cause}\label{sec:production}\label{sec:attribution}

\revs{We first study how collision failure relates to temporal compression. Temporal compression sets the number of frames per group, and together with the speed of the ball and the frame rate, this number sets how far the ball moves across one group of frames. We express this distance relative to the size of the ball as the \emph{relative displacement}:}
\begin{equation}
  \rho = v_n\,k\Delta/\ell,
  \label{eq:lambda}
\end{equation}
\revs{where $v_n$ is the collision speed normal to the surface, or the relative speed of two balls, $k$ the number of frames per group, $\Delta$ the frame interval and $\ell$ the diameter of the ball. \cref{fig:law}a shows the collision failure of our generators against $\rho$ for each number of frames per group and frame rate. Averaged over all seeds, collision failure rises with the number of frames per group, from 30.2\% with 1 frame to 36.6\%, 45.0\% and 66.0\% with 2, 4 and 8, and for each number of frames per group it rises with $\rho$ over most of its range.
 In \cref{fig:law}a, the generators of Wan2.2-TI2V-5B and LTX-Video fail all 224 collisions across their range of $\rho$, mostly as missing collisions (149 and 181 of 224). \cref{fig:law}b shows that $\rho$, computed before generation, predicts collision failure when the number of frames per group, the frame rate or the speed is held out from the prediction, better than the group duration does in all 14 seed and fold pairs. A control that keeps one latent per frame but gives the generator as many tokens as 8 frames per group fails 38.5\%, against 29.9\% with 1 frame and 64.1\% with 8 on the same seed, so three quarters of the added failure come from grouping frames in time. The relative displacement thus sets how far the ball moves within a group, which leaves open where in the group the collision falls.}%

\begin{figure*}[t]
  \centering
  \includegraphics[width=\textwidth]{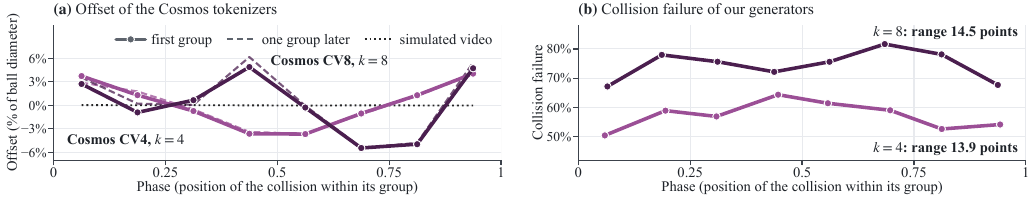}
  \caption{\textbf{Effect of the phase.} (a) The offset of the tokenizer about its mean varies with the phase and repeats one group later. (b) The collision failure of our generators varies with the phase.}
\label{fig:period}\colfit{-1.2pt}
\end{figure*}

\revs{We next study how collision failure relates to the position of the collision within its group, which we call the \emph{phase}. At fixed $k$ and $\rho$, the copies shifted one frame at a time change only this position. At the tokenizer stage, \cref{fig:period}a shows, for the two Cosmos tokenizers, the gap between the ball and the surface at the collision, the \emph{offset}, against the phase in the first group and one group later. The offset varies with the phase, and the curve one group later follows that of the first group. Offsets one group apart also agree more closely than offsets at other distances in each of the eight tokenizers, also at twice the resolution, and on up to 21 real videos of bouncing balls from Morpheus~\cite{tragoudaras2026morpheus}, the offset repeats with a period that divides the group ($p\le0.026$). This recurrence is expected from how a tokenizer treats shifts in time, since a tokenizer that shifts its output by a whole group when its input shifts by a whole group makes the same error at the collision before and after the shift. At the generation stage, \cref{fig:period}b shows the collision failure of our generators against the phase, with 4 and 8 frames per group. Collision failure varies by 13.9 and 14.5 points with this position, whereas a shift by a whole group changes it by 2.2 and 0.5 points. Failure thus depends on the phase and repeats after one group.}\looseness=-1

\revs{Finally, we study at which stage collision failure arises. Comparing the two stages separates the failure caused by the tokenizer from that added by the generator. At the tokenizer stage, \cref{tab:clockcard} lists the PSNR, the collision failure and the geometric failure of the videos decoded by the eight existing tokenizers. The decoded videos reach 39.8 to 53.6\,dB PSNR, above the 31.6 to 36.7\,dB we measure for the same tokenizers on UCF-101~\cite{soomro2012ucf101}, yet up to 58.3\% of their collisions fail, while geometric failure stays at or below 15.2\%. In six of the eight tokenizers, a wrong rebound speed accounts for 61\% to 92\% of the measured failures, and the other two, Wan2.2 and HunyuanVideo, fail at most 4.7\% of collisions. Since the median error of the offset stays below one pixel, PSNR barely registers these failures. At $512{\times}896$ instead of $256{\times}448$ pixels, with the ball's pixel size and $\rho$ matched, the tokenizers fail 9.6\% against 8.1\% (supplementary material). Our own tokenizer with 8 frames per group shows the same pattern, with a collision failure of 40.4\% but a geometric failure of 5.8\%. At the generation stage, collision failure remains high where the tokenizer is accurate. At 60 frames per second, our tokenizer keeps 97.5\% of the collisions accurate, yet our generator fails 54.2\% with 8 frames per group against 28.2\% with 1. The tokenizer thus mainly breaks the rebound speed and the generator adds failure on long groups, so the two targets for repair are the grouping of frames around the collision time and the rebound speed in the tokenizer.}\looseness=-1
\section{Method}\label{sec:method}\label{sec:formulation}

As shown in \cref{fig:alc_framework}, our method, Adaptive Latent Clock (\ALC), keeps the generation pipeline of the models analyzed above and changes two of its parts. Given the first frame, the encoder of the tokenizer encodes it into a latent, the generator produces the latents of the following groups of frames, and the decoder of the tokenizer decodes all latents into frames. In existing models, the frames are sampled at evenly spaced times, so each group of frames spans the same duration. Since collision failure rises with the distance that the ball moves across one group, the first part, Frame Timing, shortens the groups around each collision. Before generation, it sets the time of each frame, the generator reads these frame times in its time encoding, and the decoded frames can be interpolated back to evenly spaced times. Since the tokenizer mainly breaks the rebound speed, the second part, Rebound Supervision, supervises the speed of the ball around the collision when the tokenizer is trained, and the generator is then trained on the latents of this tokenizer.\looseness=-1

\subsection{Frame Timing}\label{sec:alc}\label{sec:approach}\label{sec:allocation}\label{sec:generation}

Our analysis shows that collision failure rises with the relative displacement, the displacement of the ball across one group of frames relative to the size of the ball. To reduce the relative displacement at each collision at fixed temporal compression, Frame Timing shortens the groups around each collision by sampling frames more densely there and more sparsely elsewhere. It proceeds in three steps: it estimates the collisions, allocates the frame times, and lets the generator read these frame times.\looseness=-1

To decide where to concentrate frames, Frame Timing first estimates the time and the relative displacement of each collision, the quantity that our analysis shows to predict collision failure before generation. When the initial state of the scene is known, both follow from kinematics. Otherwise, the same generator produces a fast preview with few sampling steps and evenly spaced frame times, in which we track the ball and fit its approach to the surface under the gravity of the scene. When tracking fails, Frame Timing keeps evenly spaced frame times. We then divide the video into time bins of equal length and assign to each bin the largest relative displacement among the collisions near it.\looseness=-1

Since the position of a collision within its group is unknown before generation, Frame Timing allocates the frames to minimize the largest relative displacement in the video. Let $\rho_j$ be the relative displacement assigned to bin $j$, and $w_j$ the density of frames in bin $j$ relative to evenly spaced frame times. A density $w_j$ shortens the groups in bin $j$ by a factor of $w_j$ and reduces their relative displacement to $\rho_j/w_j$, so Frame Timing chooses the densities by
\begin{equation}
  \min_{w}\,\max_j\ \rho_j/w_j,
  \label{eq:lp}
\end{equation}
with the average density fixed at one, which keeps the duration and the number of frames of the video, with each density within fixed bounds and with adjacent densities within a fixed ratio. Dividing the cumulative density evenly then gives the time of each frame, and each group of $k$ consecutive frames spans less time in dense bins.\looseness=-1

Since collision failure also arises at the generation stage, the generator has to read these frame times. A generator that reads only the index of each latent would take a dense group of frames for slow motion, so we replace the index of each latent in its temporal position encoding by the time of the last frame of its group, through a rotary time encoding~\cite{su2024roformer,wang2026bullettime}, and add an embedding of the duration of each group. To start fine-tuning from the generator with evenly spaced frame times, we initialize this embedding to zero. Fine-tuning mixes evenly spaced frame times with those of Frame Timing, computed from the simulated collisions with added errors in the estimated collision times. The decoder then produces the frames at their frame times.\looseness=-1

\subsection{Rebound Supervision}\label{sec:velocity}

Our analysis shows that the tokenizer mainly breaks the rebound speed, while PSNR barely registers this error. Rebound Supervision therefore corrects the rebound speed in the tokenizer. Existing reconstruction training does not correct it. A collision affects only a few frames, and a reconstruction loss averaged over all pixels and frames gives these frames little weight. Moreover, a decoder that cannot tell where the collision falls within its group blends the outputs for the possible positions, which rounds off the rebound and mixes the speeds before and after the collision. Rebound Supervision thus directly supervises the motion of the ball around the collision when the tokenizer is trained.\looseness=-1

Let $p_i$ be the center of the ball in frame $i$ of the simulated video, $\hat p_i$ the center of the ball in frame $i$ of the reconstructed video, computed as the centroid of the soft coverage map of the ball, and $\mathcal C$ the frames in which the ball is within four pixels of the surface. Rebound Supervision compares the displacement of the ball between consecutive frames in the reconstructed and the simulated video over $\mathcal C$:
\begin{equation}
  \mathcal{L}_{\rm speed}=\frac{1}{2|\mathcal C|}\sum_{i\in\mathcal C}\big\|(\hat p_{i+1}-\hat p_i)-(p_{i+1}-p_i)\big\|_1.
  \label{eq:vel}
\end{equation}
We add $\mathcal{L}_{\rm speed}$ to the reconstruction loss of the tokenizer with a weight that decreases to zero by the end of training, and the generator is then trained on the latents of this tokenizer. At inference, $\mathcal{L}_{\rm speed}$ is no longer computed, and the trained encoder and decoder are used directly.\looseness=-1

\section{Experiments}\label{sec:experiments}

\subsection{Experimental Setup}\label{sec:setup}

Our tokenizer is a causal 3D autoencoder with 21.7M parameters at $k{=}8$ that represents each group of $k$ frames with $4k$ channels, which gives every $k$ the same latent size per second, and trains with L1, LPIPS~\cite{zhang2018lpips} and KL losses. Our generator is a rectified-flow diffusion transformer that generates latents from the first-frame latent and the scene parameters~\cite{liu2023rectified,lipman2023flow,peebles2023dit}. For every $k$ we train two seeds on 960k samples and a third seed at $k\in\{1,8\}$. The time encoding of Frame Timing is fine-tuned for 20{,}000 steps, and at $k{=}8$ also for 60{,}000 steps with the learning rate annealed to zero over the last 10{,}000. Generators with 675M parameters, of DiT-XL size, and 1.8B parameters follow the same recipe on one seed. Frame Timing solves \cref{eq:lp} as a linear program with HiGHS~\cite{huangfu2018highs}, in 4.5\,ms per video.\looseness=-1

\Uni{} is our generator with evenly spaced frame times, and \ClockU{} fine-tunes it with the time encoding of Frame Timing and keeps these times. \ALC{} runs the fine-tuned generator with its own frame times, so its difference from \ClockU{} isolates frame allocation. Unless Rebound Supervision is named, \ALC{} runs only Frame Timing and decodes with the \emph{plain} tokenizer of \ClockU{}, trained without Rebound Supervision. By default, \ALC{} reads the collision from the known initial state. To compare with existing work, we keep our frame allocation and the 12-step preview and replace the signal that places the frames. We take one adaptive method per signal, motion for DLFR-Gen~\cite{yuan2025dlfrgen}, the closest prior method, quality for AdapTok~\cite{li2026adaptok}, uniqueness for UniComp~\cite{yuan2026unicomp} and surprise for SURGE~\cite{tang2026surge}, and add the path kinks of Ramer--Douglas--Peucker (RDP)~\cite{ramer1972iterative,douglas1973algorithms}. We reimplement these signals in our frame allocation.\looseness=-1

All results use the main test set of 5{,}382 videos at 30\,fps, shared noise and a 50-step sampler, by default with one seed at $k{=}1$ and two at ${k\in\{4,8\}}$. The tables split failures into violated, missing and undecided collisions, the last being unmeasurable. Since \ALC{} moves the frames, the tables also check the speed before the collision. The \emph{speed ratio} is its median over the simulated value, in band is the share of videos with a ratio from 0.8 to 1.25, and failure with speed also counts the videos outside this band. We report seed means, standard deviations and paired bootstrap 95\% intervals over videos. The supplementary material gives the full algorithm, hyperparameters, thresholds and per-seed results.\looseness=-1

\vspace{-4pt}%
\subsection{Main Results}\label{sec:main_results}

\begin{table*}[t]
\centering\footnotesize
\renewcommand{\sd}[1]{{\fontsize{5.4}{6}\selectfont\color{black!45}\,$\pm$#1}}%
\renewcommand{\gain}[1]{\textcolor{lcGain}{#1}}%
\setlength{\tabcolsep}{3.4pt}\renewcommand{\arraystretch}{0.89}%
\lcfit{\textwidth}{13}{%
\begin{tabular}{@{}ll cc c ccc ccc cc@{}}
\toprulex
\multirow{2}{*}{\textbf{System}} & \multirow{2}{*}{\revl{\textbf{Frame times}}} & \multicolumn{3}{c}{\textbf{Collision failure}} & \multicolumn{3}{c}{\textbf{By outcome}} & \multicolumn{3}{c}{\revl{\textbf{Collision speed}}} & \multicolumn{2}{c}{\textbf{Cost}} \\
\cmidrule(lr){3-5}\cmidrule(lr){6-8}\cmidrule(lr){9-11}\cmidrule(l){12-13}
 & & geometric\dn & rebound\dn & \textbf{all}\dn & violated\dn & missing\dn & undecided\dn & failure\dn & ratio & in band\up & evals & time\,(s)\dn \\
\midrule
\tgroup{13}{\revs{Both Modules, 33M Generator, $k{=}8$, 1.92M Samples, \revc{60k} Fine-Tuning Steps}}
\revs{\Uni{}} & \revs{evenly spaced} & \revc{19.6\sd{1.7}} & \revc{25.1\sd{0.8}} & \revc{39.8\sd{0.6}} & \revc{35.7\sd{0.1}} & \revc{0.9\sd{0.3}} & \revc{3.2\sd{0.3}} & \revc{57.5\sd{0.5}} & \revc{0.86} & \revc{53.7\sd{0.4}} & 50 & 0.3 \\
\revs{\ClockU} & \revs{evenly spaced} & \revc{\snd{6.6}\sd{0.4}} & \revc{\snd{23.0}\sd{0.5}} & \revc{\snd{28.1}\sd{0.3}} & \revc{\snd{26.8}\sd{0.3}} & \revc{\snd{0.1}\sd{0.0}} & \revc{\snd{1.2}\sd{0.1}} & \revc{\snd{47.0}\sd{0.8}} & \revc{\snd{0.87}} & \revc{\snd{60.2}\sd{0.9}} & 50 & 0.6 \\
\rowcolor{lcOursRow}\revs{\textbf{\ALC{} (ours)}} & \revs{initial state} & \revc{\bst{2.8}\sd{0.3}} & \revc{\bst{2.1}\sd{0.2}} & \revc{\bst{4.8}\sd{0.4}} & \revc{\bst{4.3}\sd{0.3}} & \revc{\bst{0.0}\sd{0.0}} & \revc{\bst{0.5}\sd{0.0}} & \revc{\bst{8.9}\sd{0.4}} & \revc{\bst{1.00}} & \revc{\bst{94.9}\sd{0.3}} & 50 & 0.6 \\
$\Delta$ & & \revc{\gain{$-$3.7}} & \revc{\gain{$-$20.9}} & \revc{\gain{$-$23.2}} & \revc{\gain{$-$22.5}} & \revc{\gain{$-$0.1}} & \revc{\gain{$-$0.7}} & \revc{\gain{$-$38.1}} & \revc{\gain{+0.13}} & \revc{\gain{+34.7}} & & \\
\tgroup{13}{33M Generator, $k{=}4$}
\Uni{} & evenly spaced & \revc{27.4\sd{1.4}} & \revc{23.8\sd{2.1}} & \revc{44.6\sd{2.6}} & \revc{37.8\sd{2.0}} & \revc{3.3\sd{0.1}} & \revc{3.6\sd{0.5}} & \revc{61.7\sd{2.1}} & \revc{0.85} & \revc{52.3\sd{1.3}} & 50 & 0.3 \\
\ClockU & evenly spaced & \revc{\snd{18.5}\sd{1.5}} & \revc{\snd{20.6}\sd{0.6}} & \revc{\snd{35.3}\sd{1.7}} & \revc{\snd{31.4}\sd{1.3}} & \revc{\snd{1.6}\sd{0.0}} & \revc{\snd{2.3}\sd{0.4}} & \revc{\snd{54.5}\sd{2.2}} & \revc{\snd{0.86}} & \revc{\snd{56.9}\sd{1.7}} & 50 & 0.6 \\
\rowcolor{lcOursRow}\textbf{\ALC{} (ours)} & initial state & \revc{\bst{10.7}\sd{0.8}} & \revc{\bst{9.5}\sd{0.8}} & \revc{\bst{19.2}\sd{1.5}} & \revc{\bst{18.5}\sd{1.4}} & \revc{\bst{0.1}\sd{0.0}} & \revc{\bst{0.5}\sd{0.0}} & \revc{\bst{33.3}\sd{2.2}} & \revc{\bst{0.94}} & \revc{\bst{77.3}\sd{1.7}} & 50 & 0.6 \\
$\Delta$ & & \revc{\gain{$-$7.7}} & \revc{\gain{$-$11.2}} & \revc{\gain{$-$16.1}} & \revc{\gain{$-$12.9}} & \revc{\gain{$-$1.5}} & \revc{\gain{$-$1.8}} & \revc{\gain{$-$21.2}} & \revc{\gain{+0.09}} & \revc{\gain{+20.4}} & & \\
\tgroup{13}{33M Generator, $k{=}8$}
\Uni{} & evenly spaced & 41.9\sd{1.6} & 40.6\sd{1.7} & 65.5\sd{1.9} & 52.0\sd{1.6} & 3.4\sd{0.6} & 10.1\sd{0.3} & 79.2\sd{1.0} & 0.77 & 33.6\sd{0.6} & 50 & 0.3 \\
\ClockU & evenly spaced & \snd{31.7}\sd{0.7} & \snd{38.5}\sd{0.3} & \snd{58.0}\sd{0.2} & \snd{48.5}\sd{0.1} & \snd{1.9}\sd{0.2} & \snd{7.6}\sd{0.0} & \snd{74.1}\sd{0.1} & \snd{0.78} & \snd{37.6}\sd{0.5} & 50 & 0.6 \\
\rowcolor{lcOursRow}\textbf{\ALC{} (ours)} & initial state & \revp{\bst{14.6}\sd{0.0}} & \revp{\bst{20.7}\sd{1.4}} & \revp{\bst{32.3}\sd{1.2}} & \revp{\bst{30.6}\sd{1.0}} & \bst{0.2}\sd{0.1} & \revp{\bst{1.4}\sd{0.3}} & \revp{\bst{47.7}\sd{0.9}} & \revp{\bst{0.95}} & \revp{\bst{66.3}\sd{1.4}} & 50 & 0.6 \\
$\Delta$ & & \revp{\gain{$-$17.1}} & \revp{\gain{$-$17.8}} & \revp{\gain{$-$25.7}} & \revp{\gain{$-$17.9}} & \gain{$-$1.7} & \revp{\gain{$-$6.2}} & \revp{\gain{$-$26.4}} & \revp{\gain{+0.17}} & \revp{\gain{+28.7}} & & \\
\tgroup{13}{675M Generator, $k{=}8$}
\revm{\Uni{}} & \revm{evenly spaced} & \revm{28.9}\revr{\sd{0.6}} & \revm{38.9}\revr{\sd{0.8}} & \revm{56.6}\revr{\sd{0.7}} & \revm{46.7}\revr{\sd{0.7}} & \revm{2.8}\revr{\sd{0.2}} & \revm{7.1}\revr{\sd{0.3}} & \revm{70.5}\revr{\sd{0.6}} & \revm{0.80} & \revm{41.2}\revr{\sd{0.7}} & \revm{50} & \revr{0.7} \\
\revm{\ClockU} & \revm{evenly spaced} & \revm{\snd{22.6}}\revr{\sd{0.6}} & \revm{\snd{38.2}}\revr{\sd{0.7}} & \revm{\snd{52.2}}\revr{\sd{0.6}} & \revm{\snd{45.0}}\revr{\sd{0.6}} & \revm{\snd{1.4}}\revr{\sd{0.2}} & \revm{\snd{5.8}}\revr{\sd{0.3}} & \revm{\snd{66.6}}\revr{\sd{0.6}} & \revm{\snd{0.81}} & \revm{\snd{43.8}}\revr{\sd{0.7}} & \revm{50} & \revr{1.4} \\
\rowcolor{lcOursRow}\revm{\textbf{\ALC{} (ours)}} & \revm{initial state} & \revm{\bst{6.9}}\revr{\sd{0.3}} & \revm{\bst{11.8}}\revr{\sd{0.4}} & \revm{\bst{17.8}}\revr{\sd{0.5}} & \revm{\bst{17.2}}\revr{\sd{0.5}} & \revm{\bst{0.1}}\revr{\sd{0.0}} & \revm{\bst{0.5}}\revr{\sd{0.1}} & \revm{\bst{29.9}}\revr{\sd{0.6}} & \revm{\bst{0.98}} & \revm{\bst{80.0}}\revr{\sd{0.6}} & \revm{50} & \revr{1.4} \\
$\Delta$ & & \revm{\gain{$-$15.7}} & \revm{\gain{$-$26.5}} & \revm{\gain{$-$34.4}} & \revm{\gain{$-$27.8}} & \revm{\gain{$-$1.3}} & \revm{\gain{$-$5.3}} & \revm{\gain{$-$36.7}} & \revm{\gain{+0.18}} & \revm{\gain{+36.2}} & & \\
\tgroup{13}{\revk{1.8B Generator, $k{=}8$}}
\revk{\Uni{}} & \revk{evenly spaced} & \revk{29.4}\revr{\sd{0.6}} & \revk{41.5}\revr{\sd{0.8}} & \revk{58.7}\revr{\sd{0.6}} & \revk{48.0}\revr{\sd{0.7}} & \revk{3.5}\revr{\sd{0.3}} & \revk{7.2}\revr{\sd{0.3}} & \revk{71.4}\revr{\sd{0.6}} & \revk{0.81} & \revk{41.5}\revr{\sd{0.7}} & \revk{50} & \revr{1.0} \\
\revk{\ClockU} & \revk{evenly spaced} & \revk{\snd{21.6}}\revr{\sd{0.6}} & \revk{\snd{38.0}}\revr{\sd{0.7}} & \revk{\snd{51.4}}\revr{\sd{0.7}} & \revk{\snd{44.0}}\revr{\sd{0.7}} & \revk{\snd{2.0}}\revr{\sd{0.2}} & \revk{\snd{5.5}}\revr{\sd{0.3}} & \revk{\snd{66.0}}\revr{\sd{0.6}} & \revk{\snd{0.82}} & \revk{\snd{45.2}}\revr{\sd{0.7}} & \revk{50} & \revr{1.8} \\
\rowcolor{lcOursRow}\revk{\textbf{\ALC{} (ours)}} & \revk{initial state} & \revk{\bst{5.6}}\revr{\sd{0.3}} & \revk{\bst{11.0}}\revr{\sd{0.4}} & \revk{\bst{15.9}}\revr{\sd{0.5}} & \revk{\bst{15.4}}\revr{\sd{0.5}} & \revk{\bst{0.1}}\revr{\sd{0.0}} & \revk{\bst{0.4}}\revr{\sd{0.1}} & \revk{\bst{27.7}}\revr{\sd{0.6}} & \revk{\bst{0.99}} & \revk{\bst{80.7}}\revr{\sd{0.5}} & \revk{50} & \revr{1.8} \\
$\Delta$ &  & \revk{\gain{$-$16.1}} & \revk{\gain{$-$27.0}} & \revk{\gain{$-$35.5}} & \revk{\gain{$-$28.5}} & \revk{\gain{$-$1.9}} & \revk{\gain{$-$5.1}} & \revk{\gain{$-$38.3}} & \revk{\gain{+0.17}} & \revk{\gain{+35.5}} &  &  \\
\tgroup{13}{\revi{Allocation Signals Adapted from Adaptive Methods, 33M Generator, $k{=}8$}}
AdapTok~\cite{li2026adaptok}\venue{quality} & preview & 56.8\sd{0.7} & \revt{46.0\sd{2.2}} & \revt{76.6\sd{1.4}} & \revt{42.5\sd{1.2}} & \revt{11.3\sd{0.5}} & \revt{22.8\sd{0.7}} & 88.3\sd{1.2} & 0.67 & \revt{19.1\sd{1.3}} & 62 & 0.8 \\
DLFR-Gen~\cite{yuan2025dlfrgen}\venue{motion} & preview & 59.3\sd{1.5} & \revt{40.6\sd{0.9}} & 75.8\sd{0.6} & \revt{37.6\sd{0.5}} & \revt{14.5\sd{0.4}} & \revt{23.7\sd{1.5}} & \revt{85.1\sd{0.7}} & \revp{\snd{0.93}} & \revt{29.2\sd{1.9}} & 62 & 0.8 \\
UniComp~\cite{yuan2026unicomp}\venue{uniqueness} & preview & \revt{44.7\sd{0.0}} & \revt{41.9\sd{0.1}} & \revt{67.9\sd{0.0}} & \revt{48.7\sd{0.1}} & \revt{4.1\sd{0.7}} & \revt{15.1\sd{0.7}} & \revt{83.5\sd{0.3}} & \revt{0.71} & \revt{25.7\sd{0.6}} & 62 & 0.8 \\
SURGE~\cite{tang2026surge}\venue{surprise} & preview & 41.1\sd{2.1} & \revt{34.9\sd{2.3}} & 61.6\sd{2.7} & \revt{42.0\sd{2.2}} & 5.8\sd{0.4} & 13.8\sd{0.8} & \revt{78.6\sd{1.1}} & 0.77 & \revt{32.0\sd{0.5}} & 62 & 0.8 \\
RDP~\cite{ramer1972iterative,douglas1973algorithms}\venue{kinks} & preview & \revp{\snd{21.4}\sd{1.4}} & \revp{\snd{24.4}\sd{1.3}} & \revp{\snd{40.6}\sd{0.0}} & \revp{\snd{36.0}\sd{0.6}} & \revp{\snd{2.4}\sd{0.6}} & \revp{\snd{2.2}\sd{0.0}} & \revp{\snd{57.7}\sd{0.2}} & \revp{\snd{0.93}} & \revp{\snd{57.3}\sd{1.1}} & 62 & 0.8 \\
\rowcolor{lcOursRow}\textbf{\ALC{} (ours)}\venue{\revl{collision}} & preview & \revp{\bst{16.6}\sd{0.1}} & \revp{\bst{19.8}\sd{1.6}} & \revp{\bst{33.0}\sd{1.5}} & \revp{\bst{29.9}\sd{1.6}} & \revp{\bst{1.3}\sd{0.3}} & \revp{\bst{1.8}\sd{0.2}} & \revp{\bst{49.1}\sd{1.6}} & \revp{\bst{0.95}} & \revp{\bst{64.0}\sd{1.6}} & 62 & 1.0 \\
\bottomrulex
\end{tabular}}
\caption{\revl{\textbf{\ALC{} on our generators.} Test collision failure (\%). \ALC{} in the first group adds Rebound Supervision. $\Delta$ is \ALC{} minus \ClockU{} from unrounded values, bold and underline the best two per group, grey the deviation over seeds or videos.}}
\label{tab:main}\colfit{-1pt}
\end{table*}

The first group of \cref{tab:main} runs both modules with generators trained on 1.92M samples, \revc{on three seeds}, and \ALC{} lowers collision failure at $k{=}8$ from \revc{28.1\%} to \revc{4.8\%}, which the ablation studies break down by part. The other groups test Frame Timing by itself, with the same generator, tokenizer, noise, 50 network evaluations and latent count. At $k{=}8$, \ALC{} lowers collision failure from 58.0\% for \ClockU{} to 32.3\%, with 95\% intervals below $-23.2$ and $-25.2$ on the two seeds. Geometric failure falls by 17.1 points, and among the collisions that pass the geometric checks, rebound failure falls from 38.5\% to 20.7\%. At $k{=}4$, failure falls from 35.3\% to 19.2\%, and at the $k{=}4$ setting the seed-1 gain grows from 9.1 points at $k{=}1$ to 22.6 at $k{=}8$. With 7 latents per video at $k{=}8$, \ALC{} thus fails less than \ClockU{} with 13 at $k{=}4$. The gain holds in every quarter of the phase, so it stems from the shorter groups, not from a favorable phase.\looseness=-1

\ALC{} also places frames better than the signals of adaptive methods. At the bottom of \cref{tab:main}, it fails 28.6 to 43.6 points less than the four adaptive signals, and 6.5 and 8.6 points less than RDP on the two seeds, all 95\% intervals below zero. When every signal also decodes with the tokenizer trained with Rebound Supervision, \ALC{} leads all five by 7.2 to 40.1 points. These signals follow appearance and motion across the video, so their shortest group lies a median 230 to 660\,ms from the collision, against 38 to 50\,ms for \ALC{} and RDP.\looseness=-1

\begin{figure}[t]
  \centering
  \includegraphics[width=\columnwidth]{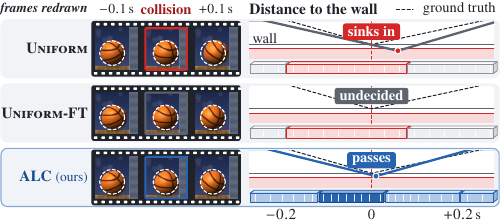}
  \caption{\revm{\textbf{One \revf{wall} collision \revl{under the three systems}\revf{.}}} \revl{The ball is redrawn at its measured position in each frame.}}
\label{fig:wall_case}\colfit{-6pt}%
\end{figure}

The gain grows with training and holds with scale. After 60{,}000 fine-tuning steps (supplementary material), \ALC{} fails 17.1\% against 47.6\% for \ClockU{}, $-30.5$ points with intervals below $-28.7$ and $-29.3$. The 675M and 1.8B generators still fail 52.2\% and 51.4\% as \ClockU{} and 17.8\% and 15.9\% with \ALC{}. Growing the generator 55-fold to 1.8B parameters thus lowers the failure of \ClockU{} by 6.6 points, whereas the gain of \ALC{} widens from 25.7 points at 33M to 34.4 at 675M and 35.5 at 1.8B. On every trained generator, \ALC{} removes 37\% to 74\% of the geometric failures of \ClockU{}. \Cref{fig:wall_case} follows one such collision, whose group shortens from 267 to 150\,ms.\looseness=-1

\begin{table}[t]
\revmon
\centering\footnotesize
\setlength{\tabcolsep}{6pt}\renewcommand{\arraystretch}{0.84}%
\lcfit{\columnwidth}{3}{%
\begin{tabular}{@{}lcc@{}}
\toprulex
\textbf{\revf{Failed first collisions}} & Evenly spaced & \ALC{} \\
\midrule
Frozen tokenizer & \revf{152} & \revf{\bst{4}} \\
Regeneration, $s{=}0.1$ & \revf{121} & \revf{\bst{30}} \\
Regeneration with LoRA, $s{=}0.2$ & \revf{132} & \revf{\bst{48}} \\
\revw{Wan2.2-TI2V-5B} & \revw{22} & \revw{\bst{2}} \\
\bottomrulex
\end{tabular}}
\caption{\revk{\textbf{\ALC{} on \revl{existing models}.} \revw{The first three rows use LTX-Video, and the LoRA row fine-tunes its generator.}}}
\label{tab:ltx}\colfit{2pt}%
\end{table}

\cref{tab:ltx} applies \ALC{} to existing models.\label{sec:ltx} Without any training, we encode and decode 300 further drops of 2\,s with the eight tokenizers of \cref{tab:clockcard}. The frame times of \ALC{} lower their geometric failures from 657 to 134 of the 1{,}829 collisions that pass in the simulated frames, and from 152 to 4 of 228 for the $k{=}8$ tokenizer of LTX-Video. On our 240\,fps footage of a table-tennis ball, played at half speed, all eight fail 556 of 3{,}730 collision windows with evenly spaced frame times and 141 with those of \ALC{} (supplementary material). Regenerating the 466 windows of LTX-Video that pass in the filmed frames at strength 0.2, \ALC{} passes 143 collisions that evenly spaced frame times fail, against 52 the other way. With the generator of LTX-Video unchanged, we also regenerate 200 new videos by video-to-video denoising at strength 0.1. Evenly spaced frame times and those of \ALC{} share 49 frames, noise and prompt, and the first collision is located in the input video. Scored on the generated frames with the decoder of LTX-Video, \ALC{} fails 30 of the 152 first collisions whose input video passes every check, against 121 with evenly spaced frame times. At strengths 0.2 and 0.4, \ALC{} passes 60 and 24 collisions that evenly spaced frame times fail, against 8 and 9 the other way. Frame allocation thus transfers to a pretrained generator. A rank-16 LoRA that reads the frame times through the rotary encoding raises these counts to 87 against 3 and 65 against 9. On Wan2.2-TI2V-5B, the same regeneration with the collision read from the initial state fails 2 of 150 first collisions with \ALC{}, against 22 with evenly spaced frame times.\looseness=-1

With three collisions per video, 109 of 200 new videos pass every scorable collision with \ALC{}, against 64 with evenly spaced frame times and 82 with the DLFR-Gen signal. On 200 further videos, giving each collision the $\rho$ of its neighbor lowers the passing videos from 96 to 71 ($p{=}0.004$), so each collision needs its own frames.\looseness=-1

\vspace{0pt}%
\subsection{Ablation Studies}\label{sec:ablation}

\begin{table}[t]
\centering\footnotesize
\setlength{\tabcolsep}{2.6pt}\renewcommand{\arraystretch}{0.95}%
\lcfit{\columnwidth}{7}{%
\begin{tabular}{@{}lcccccc@{}}
\toprulex
\textbf{System} & geom.\dn & reb.\dn & \textbf{all}\dn & with speed\dn & ratio & in band\up \\
\midrule
\tgroup{7}{\revl{Frame Timing and Rebound Supervision}\revf{, 1.92M Samples}}
\rowcolor{lcOursRow}\revm{Full} & \revc{2.8} & \revc{2.1} & \revc{\bst{4.8}} & \revc{\bst{8.9}} & \revc{\bst{1.00}} & \revc{94.9} \\
\revl{w/o Rebound Supervision} & \revc{2.5} & \revc{8.4} & \revc{10.7} & \revc{21.5} & \revc{0.95} & \revc{84.8} \\
\revl{w/o Frame Timing} & \revc{5.9} & \revc{1.2} & \revc{7.0} & \revc{10.0} & \revc{0.99} & \revc{95.6} \\
\revm{w/o both} & \revc{6.6} & \revc{23.0} & \revc{28.1} & \revc{47.0} & \revc{0.87} & \revc{60.2} \\
\tgroup{7}{\revk{Reference}\revf{, 960k Samples}}
\ClockU{} & 31.7 & 38.5 & 58.0 & 74.1 & 0.78 & 37.6 \\
\rowcolor{lcOursRow}\ALC{} & \revm{14.6} & \revm{20.7} & \revm{32.3} & \revm{47.7} & \revm{0.95} & \revm{66.3} \\
\tgroup{7}{\revl{Frame Timing: Frame Allocation}}
\revm{Collision-weighted training} & 30.2 & 38.0 & 56.7 & 72.6 & 0.79 & 39.3 \\
\revl{Group durations time-reversed} & \revm{55.8} & \revm{34.2} & \revm{70.9} & \revm{85.5} & \revm{0.75} & \revm{26.1} \\
\revl{Allocation proportional to $\rho$} & \revm{21.0} & \revm{23.9} & \revm{39.9} & \revm{63.2} & \revm{1.04} & \revm{48.9} \\
\tgroup{7}{\revk{Frame Timing: Time Encoding}}
\revm{Frame indices as frame times} & \revm{21.5} & \revm{12.5} & \revm{31.3} & \revm{98.5} & \revm{1.96} & \revm{4.8} \\
\revm{\ALC{} resampled to 30\,fps} & \revm{14.9} & \revm{17.9} & \revm{30.1} & \revr{45.1} & \revm{0.98} & \revm{69.2} \\
\revm{\ClockU{} resampled to \ALC{} times} & \revr{34.3} & \revm{41.2} & \revm{61.4} & \revr{80.3} & \revm{0.72} & \revm{30.4} \\
\bottomrulex
\end{tabular}}
\captionsetup{skip=3pt}\caption{\revk{\textbf{Ablations on the test set}, in percent at $k{=}8$.}}
\label{tab:controls}\colfit{-5pt}%
\end{table}

The top group of \cref{tab:controls} removes Frame Timing, Rebound Supervision, or both.\label{sec:vae} Its rows share the setting of the first group of \cref{tab:main}. \revc{Removing Rebound Supervision raises failure from 4.8\% to 10.7\%, as rebound failure rises from 2.1\% to 8.4\% while geometric failure stays near 3\%. Removing Frame Timing raises it to 7.0\%, as geometric failure rises from 2.8\% to 5.9\% while rebound failure stays at or below 2.1\%. Removing both raises it to 28.1\%.} Each part thus repairs its own target, and their gains overlap: Frame Timing removes \revc{17.3} points on the plain tokenizer but \revc{2.2} on the tokenizer with Rebound Supervision. Without any training, the frame times of \ALC{} also lower the plain tokenizer's failure on the rendered test videos from 40.4\% to 8.0\%.\looseness=-1

For collision estimation, the preview nearly matches the initial state, 33.0\% against 32.3\%, and 5.3\% against 4.8\% with both modules. At 50 evaluations, as for \mbox{\ClockU{}}, failure changes by $-0.1$ points. A 67\,ms collision-time error keeps \ALC{} 14.7 to 26.9 points below \ClockU{}.\looseness=-1

For frame allocation, three changes show that the gain comes from placing the short groups. Collision-weighted training, giving a model with evenly spaced frame times the same exposure to collisions, gains 1.3 points over \ClockU{}. Reversing the group durations of \ALC{} in time misplaces the allocation and raises failure 12.9 points above \ClockU{}. Allocating frames in proportion to $\rho$ without \cref{eq:lp} fails 7.6 points more than \ALC{}.\looseness=-1

For the time encoding, frame indices in place of frame times run motion 1.96 times too fast, so failure with speed reaches 98.5\%, against 47.7\% for \ALC{}. The gain arises in generation. Resampling the output of \ClockU{} to the frame times of \ALC{} leaves its geometric failure at 34.3\%, whereas generating at those times lowers it to 14.6\%. Resampled to 30\,fps, \ALC{} keeps its gain at 30.1\%.\looseness=-1

\ifdefined\paramvariant\else\def\paramvariant{text}\fi
\def\pvfig{fig}\def\pvtab{tab}
\def\pvtext{text}\ifx\paramvariant\pvtext\def\scalecite{}\else\def\scalecite{As \cref{tab:scale} shows, at}\fi
For Rebound Supervision, we train a new pair of tokenizers, plain and with Rebound Supervision, for 40{,}000 steps and a new generator on each, so their rates differ from the other groups of \cref{tab:main}. Without a generator, Rebound Supervision lowers the reconstruction failure at $k{=}8$ from 23.3\% to 0.1\%, shrinks the swing of the offset over one group 16-fold and raises PSNR by 1.1\,dB. Optical-flow and speed errors near the collision fall 36\% and 24\%, and retraining the decoder of the frozen plain tokenizer lowers failure from 17.2\% to 5.6\% with Rebound Supervision. The plain latents thus encode the phase, and Rebound Supervision teaches the decoder to read it. In all six tokenizer and data settings (supplementary material), \ALC{} removes 26\% to 49\% of the failures of \ClockU{} and 40\% to 46\% of its geometric ones.\looseness=-1

\vspace{0pt}%
\subsection{\revk{Parameter Analysis}}\label{sec:parameters}

\ifx\paramvariant\pvfig
\begin{figure}[t]
  \centering
  \includegraphics[width=\columnwidth]{figures/fig6_params.pdf}
  \caption{\revk{\textbf{Parameter analysis on the validation set at $k{=}8$.} PSNR is dotted in (c), and circles mark the chosen values.}}
\label{fig:params}\colfit{-6pt}
\end{figure}
\revl{\ALC{} has two main parameters, the largest density $w_{\max}$ and the half-width of the window, and Rebound Supervision has its weight. \Cref{fig:params} varies each on the validation set. Failure stays within two points for $w_{\max}$ from 3 to 5 and half-widths from 0.06 to 0.10\,s, and rises when a cap of 2 leaves the collision too few frames or a window of 0.15\,s spreads them. Every weight of $\mathcal{L}_{\rm speed}$ removes the failure after reconstruction and the smallest keeps PSNR.}\looseness=-1
\else\ifx\paramvariant\pvtab
\input{tables/tab_params}
\revl{\ALC{} has two main parameters, the largest density $w_{\max}$ and the half-width of the window, and Rebound Supervision has its weight. \Cref{tab:params} varies each on the validation set. Failure stays within two points for $w_{\max}$ from 3 to 5 and half-widths from 0.06 to 0.10\,s, and rises when a cap of 2 leaves the collision too few frames or a window of 0.15\,s spreads them. Every weight of $\mathcal{L}_{\rm speed}$ removes the failure after reconstruction and the smallest keeps PSNR.}\looseness=-1
\else
On the validation set at $k{=}8$, the four allocation parameters, the two bounds on the density, the largest ratio between adjacent bins and the half-width around a collision, keep failure between 38.7\% and 44.6\% for largest densities of at least 2.5 and half-widths of at most 0.10\,s, and we use 0.5, 3, 1.5 and 0.07--0.08\,s. Failure is 70.2\% at a largest density of 2 and 53.9\% at 0.15\,s. Weights of $\mathcal{L}_{\rm speed}$ from 0.03 to 0.3 all lower the failure after reconstruction from 21.1\% to at most 0.3\%.\looseness=-1
\fi\fi

\vspace{0pt}%
\section{Conclusion}\label{sec:conclusion}

On LatentClock-Bench, collision failure under temporal compression follows the displacement of the ball within one group relative to its size, repeats with a period of one group, and persists at high PSNR and with a 55-fold larger generator. The times at which the frames of each group are sampled set this displacement. \ALC{} adapts these times with Frame Timing at fixed temporal compression and corrects the rebound with Rebound Supervision, lowering collision failure from \revc{28.1\%} to \revc{4.8\%} on LatentClock-Bench and from 121 to 30 of 152 regenerated videos on LTX-Video. Since the relative displacement is known before generation, a model can spend its fixed number of frames and latents where its collisions need them.\looseness=-1

\newpage
\section*{LLM Usage Statement}
The authors conceived the research question and hypotheses, designed every experiment and analysis, made all research
decisions, and checked the derivations and numbers against the result files. Under their direction, LLM-based assistants
(Claude, Codex, Gemini and ChatGPT) served as tools to suggest edits for polishing the text and to implement and run code. The authors take full responsibility for the content.\looseness=-1

{
    \small
    \nocite{hacohen2026ltx2,hunyuan2025video15}%
    \bibliographystyle{ieeenat_fullname}
    \bibliography{refs}
}

\clearpage
\AddToShipoutPicture*{\AtPageLowerLeft{\includegraphics[page=1,width=\paperwidth,height=\paperheight]{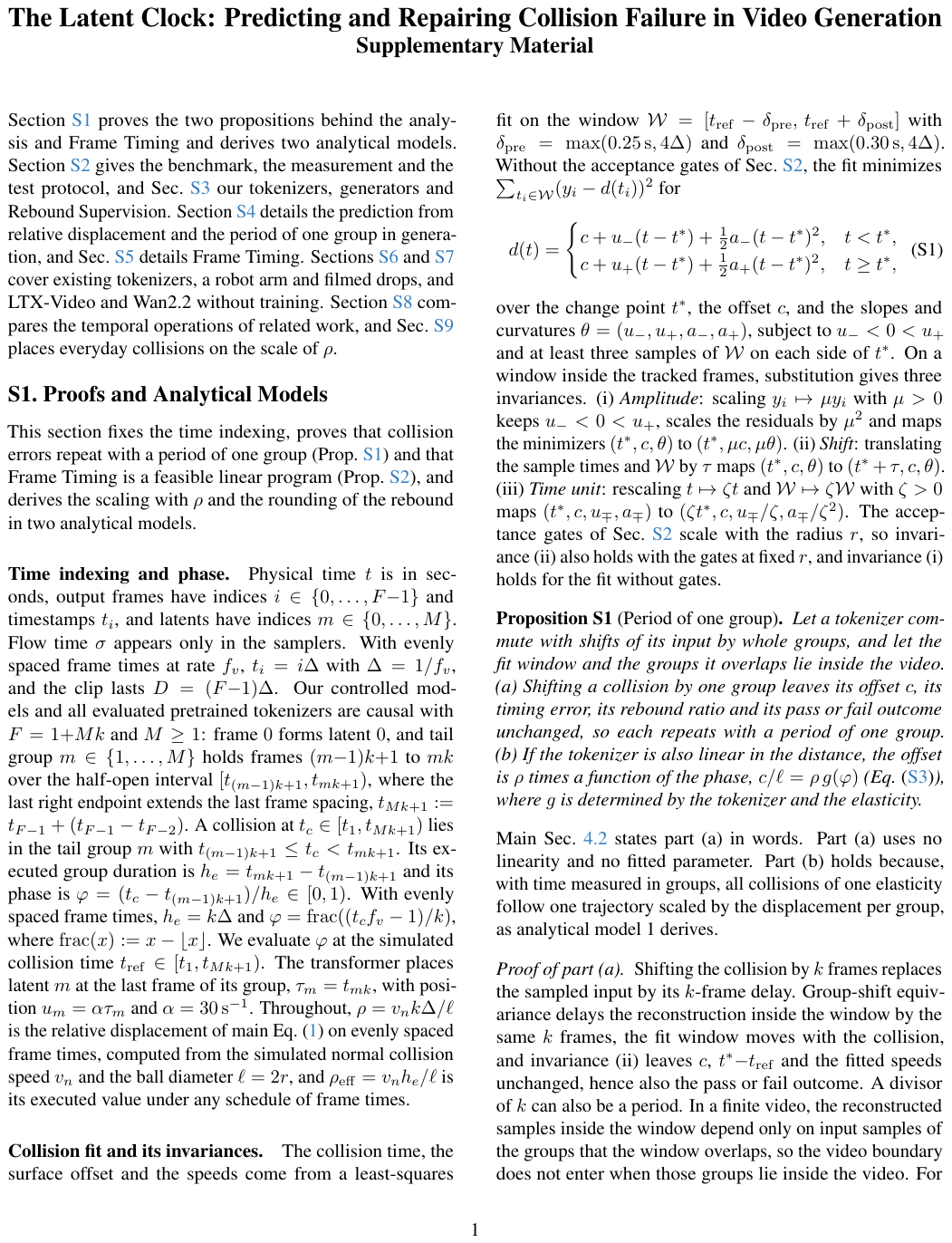}}}\thispagestyle{empty}\null\clearpage
\AddToShipoutPicture*{\AtPageLowerLeft{\includegraphics[page=2,width=\paperwidth,height=\paperheight]{supplementary.pdf}}}\thispagestyle{empty}\null\clearpage
\AddToShipoutPicture*{\AtPageLowerLeft{\includegraphics[page=3,width=\paperwidth,height=\paperheight]{supplementary.pdf}}}\thispagestyle{empty}\null\clearpage
\AddToShipoutPicture*{\AtPageLowerLeft{\includegraphics[page=4,width=\paperwidth,height=\paperheight]{supplementary.pdf}}}\thispagestyle{empty}\null\clearpage
\AddToShipoutPicture*{\AtPageLowerLeft{\includegraphics[page=5,width=\paperwidth,height=\paperheight]{supplementary.pdf}}}\thispagestyle{empty}\null\clearpage
\AddToShipoutPicture*{\AtPageLowerLeft{\includegraphics[page=6,width=\paperwidth,height=\paperheight]{supplementary.pdf}}}\thispagestyle{empty}\null\clearpage
\AddToShipoutPicture*{\AtPageLowerLeft{\includegraphics[page=7,width=\paperwidth,height=\paperheight]{supplementary.pdf}}}\thispagestyle{empty}\null\clearpage
\AddToShipoutPicture*{\AtPageLowerLeft{\includegraphics[page=8,width=\paperwidth,height=\paperheight]{supplementary.pdf}}}\thispagestyle{empty}\null\clearpage
\AddToShipoutPicture*{\AtPageLowerLeft{\includegraphics[page=9,width=\paperwidth,height=\paperheight]{supplementary.pdf}}}\thispagestyle{empty}\null\clearpage
\AddToShipoutPicture*{\AtPageLowerLeft{\includegraphics[page=10,width=\paperwidth,height=\paperheight]{supplementary.pdf}}}\thispagestyle{empty}\null\clearpage
\AddToShipoutPicture*{\AtPageLowerLeft{\includegraphics[page=11,width=\paperwidth,height=\paperheight]{supplementary.pdf}}}\thispagestyle{empty}\null\clearpage
\AddToShipoutPicture*{\AtPageLowerLeft{\includegraphics[page=12,width=\paperwidth,height=\paperheight]{supplementary.pdf}}}\thispagestyle{empty}\null\clearpage
\AddToShipoutPicture*{\AtPageLowerLeft{\includegraphics[page=13,width=\paperwidth,height=\paperheight]{supplementary.pdf}}}\thispagestyle{empty}\null\clearpage
\AddToShipoutPicture*{\AtPageLowerLeft{\includegraphics[page=14,width=\paperwidth,height=\paperheight]{supplementary.pdf}}}\thispagestyle{empty}\null\clearpage
\AddToShipoutPicture*{\AtPageLowerLeft{\includegraphics[page=15,width=\paperwidth,height=\paperheight]{supplementary.pdf}}}\thispagestyle{empty}\null\clearpage
\AddToShipoutPicture*{\AtPageLowerLeft{\includegraphics[page=16,width=\paperwidth,height=\paperheight]{supplementary.pdf}}}\thispagestyle{empty}\null\clearpage
\AddToShipoutPicture*{\AtPageLowerLeft{\includegraphics[page=17,width=\paperwidth,height=\paperheight]{supplementary.pdf}}}\thispagestyle{empty}\null\clearpage
\AddToShipoutPicture*{\AtPageLowerLeft{\includegraphics[page=18,width=\paperwidth,height=\paperheight]{supplementary.pdf}}}\thispagestyle{empty}\null\clearpage
\end{document}